\pdfoutput=1
\documentclass[11pt]{article}

\newif\ifforreview
\forreviewfalse %

\ifforreview
  \usepackage[review]{acl} %
  \renewcommand{\outauthor}{{\bf Anonymous NAACL 2021  Submission}}
\else
  \usepackage{acl}
\fi

\usepackage{etoolbox}
\usepackage{times}
\usepackage{latexsym}
\usepackage[T1]{fontenc}
\usepackage[utf8]{inputenc}
\usepackage{microtype}

\usepackage[fleqn]{amsmath}
\usepackage{url}
\usepackage{booktabs}
\usepackage{multirow}
\usepackage{graphicx}
\usepackage{amssymb}
\usepackage{bbm}

\newtoggle{comment}
\toggletrue{comment}

\newcommand\titletext{Learning Syntax from Naturally-Occurring Bracketings%
}

\newcommand{\posscite}[1]{\citeauthor{#1}'s \citeyearpar{#1}}

\newcommand{\bivec}[1]{\ensuremath{\mathbf{#1}}}

\title{\titletext}

\author{
  Tianze Shi\\
  Cornell University \\
  {\tt tianze@cs.cornell.edu} \\\And
  Ozan {\.I}rsoy\\
  Bloomberg L.P. \\
  {\tt oirsoy@bloomberg.net} \\\AND
  Igor Malioutov\\
  Bloomberg L.P. \\
  {\tt imalioutov@bloomberg.net} \\\And
  Lillian Lee\\
  Cornell University\\
  {\tt llee@cs.cornell.edu}\\
}

\date{}

\begin{document}
\maketitle

\begin{abstract}
    Naturally-occurring bracketings, such as answer fragments to natural language questions and hyperlinks on webpages,
    can reflect human syntactic intuition regarding phrasal boundaries.
    Their availability and approximate correspondence to syntax make them appealing as distant information sources to incorporate into unsupervised constituency parsing.
    But they are noisy and incomplete;
    to address this challenge,
    we develop a partial-brackets-aware structured ramp loss in learning.
    Experiments demonstrate that
    our distantly-supervised models trained on naturally-occurring bracketing data are
    more accurate in inducing syntactic structures
    than competing unsupervised systems.
    On the English WSJ corpus, our models achieve
    an unlabeled F1 score of $68.9$ for constituency parsing.\footnote{
    Our code is publicly available at
    \url{https://github.com/tzshi/nob-naacl21}.
    }
\end{abstract}

\section{Introduction}
\label{sec:intro}

\begin{figure}[t]
    \centering
    \includegraphics[width=\columnwidth]{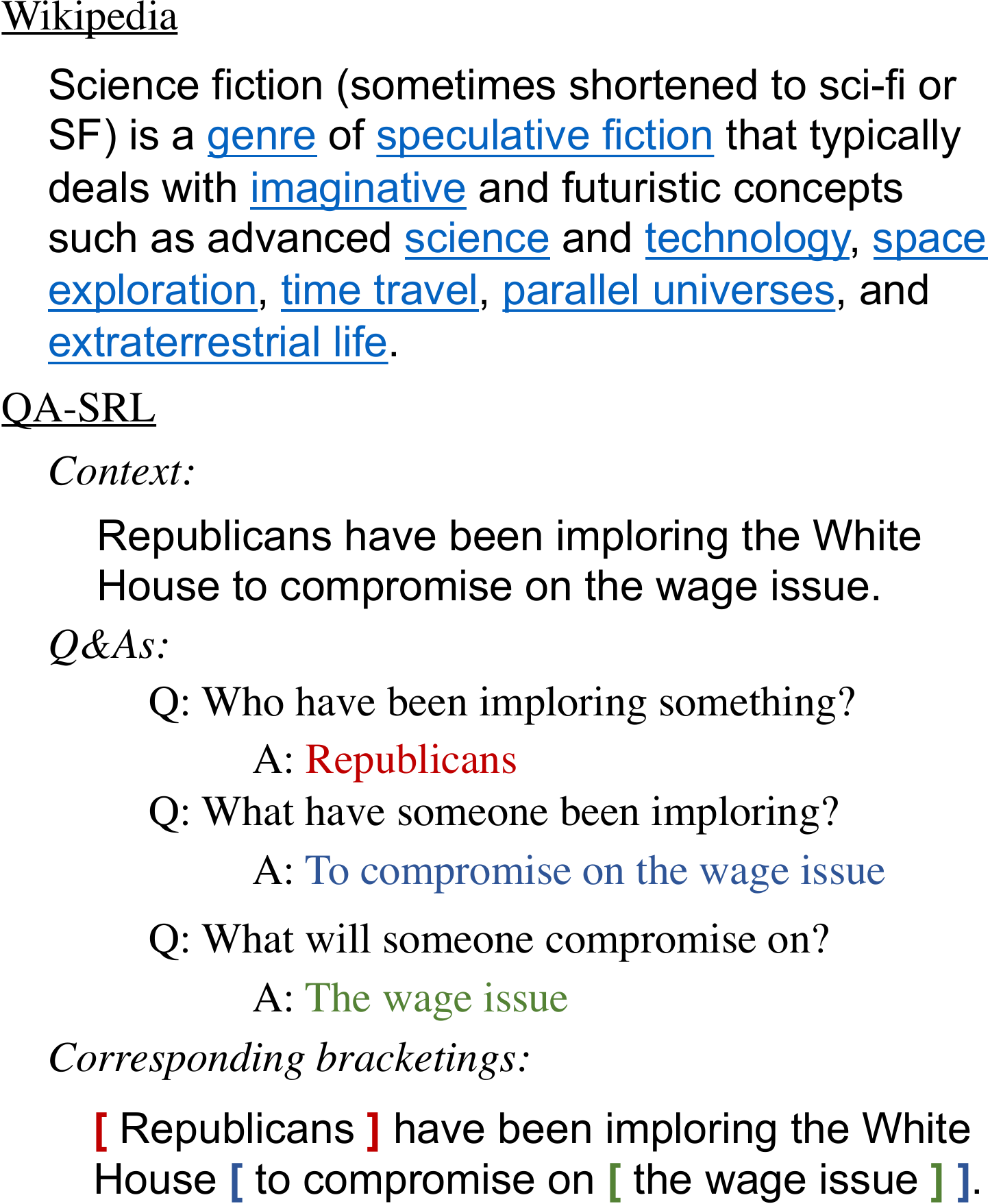}
    \caption{
        Two example types of naturally-occurring bracketings.
        Blue underlined texts in the Wikipedia sentence are hyperlinks.
        We bracket the QA-SRL sentence in matching colors according to the answers.
    }
    \label{fig:ex}
\end{figure}

Constituency is a foundational building block for phrase-structure grammars.
It captures the notion of what tokens can group together and act as a single unit.
The motivating insight behind this paper is that constituency may be reflected in mark-ups of bracketings that people provide in doing natural tasks.
We term these segments \emph{naturally-occurring bracketings}
for their lack of intended syntactic annotation.
These include, for example, the segments people pick out from sentences to refer to other Wikipedia pages or to answer semantically-oriented questions;
see Figure~\ref{fig:ex} for an illustration.

Gathering such data requires low annotation expertise and effort.
On the other hand, these data are not necessarily suitable for training parsers,
as they often contain incomplete, incorrect and sometimes conflicting bracketing information.
It is thus an empirical question
whether and how much we could learn syntax from these naturally-occurring bracketing data.

To overcome the challenge of learning from this kind of noisy data,
we propose to train discriminative constituency parsers with structured ramp loss \citep{do+08},
a technique previously adopted in machine translation \citep{gimpel-smith12}.
Specifically,
we propose two loss functions to directly penalize
predictions in conflict with available partial bracketing data,
while allowing the parsers to induce the remaining structures.

We experiment with two types of naturally-occurring bracketing data,
as illustrated in Figure~\ref{fig:ex}.
First, we consider English question-answer pairs collected for semantic role labeling (QA-SRL; \citealp{he+15b}).
The questions are designed for non-experts to specify semantic arguments of predicates in the sentences.
We observe that although no syntactic structures are explicitly asked for,
humans tend to select constituents in their answers.
Second, Wikipedia articles\footnote{We worked with articles in English.} are typically richly annotated with internal links to other articles.
These links are marked on phrasal units that refer to standalone concepts,
and similar to the QA-SRL data, they frequently coincide with syntactic constituents.

Experiment results show that naturally-occurring bracketings
across both data sources indeed help our models
induce syntactic constituency structures.
Training on the QA-SRL bracketing data achieves
an unlabeled F1 score of $68.9$
on the English WSJ corpus,
an accuracy  competitive 
with  state-of-the-art unsupervised constituency parsers
that do not utilize such distant supervision data.
We find that our proposed two loss functions have slightly different interactions with the two data sources,
and that the QA-SRL and Wikipedia data have
varying coverage of phrasal types,
leading to different error profiles.
In sum,
through this work,
(1) we demonstrate that
naturally-occurring bracketings
are helpful for inducing syntactic structures,
(2) we incorporate two new cost functions into structured ramp loss
to train parsers with noisy bracketings,
and (3) our distantly-supervised models achieve
results competitive with the state of the art
of unsupervised constituency parsing
despite training with smaller data size (QA-SRL)
or out-of-domain data (Wikipedia).

\section{Naturally-Occurring Bracketings}
\label{sec:data}

\begin{table}[t]
    \centering
    \small
    \begin{tabular}{lcc}
    \toprule
    Dataset & QA-SRL & Wikipedia \\
    \midrule
    Number of sentences & $1{,}241$ & $926{,}077$ \\
    Brackets/sentence & $6.26$  & $0.89$ \\
    \midrule
    Single word & $22.4\%$ & $35.8\%$ \\
    Constituent in reference & $55.2\%$ & $31.1\%$ \\
    Conflicting w/ reference & $11.8\%$ & $~~5.3\%$ \\
    \midrule
    SBAR & $~~2.8\%$ & $0.07\%$ \\
    NP & $36.8\%$ & $4.42\%$ \\
    VP & $~~6.3\%$ & $0.07\%$ \\
    PP & $13.3\%$ & $0.04\%$ \\
    ADJP & $~~8.6\%$ & $1.48\%$ \\
    ADVP & $30.5\%$ & $0.39\%$ \\
    \emph{Total} & $21.8\%$ & $1.91\%$ \\
    \bottomrule
    \end{tabular}
    \caption{
        Dataset statistics:
        number of bracketings per sentence (top),
        percentage of bracketing types (middle),
        and the reference phrases per label found in the natural bracketings (bottom).
        \emph{Conflicting} means the bracket crosses some reference span.
        Reference parses for Wikipedia are generated by a parser trained on PTB.
    }
    \label{tab:stats}
\end{table}

Constituents are naturally reflected in
various
human cognitive processes,
including speech production and perception \citep{garrett+66,gee-grosjean83},
reading behaviors \citep{hale01,boston+08},
punctuation marks \citep{spitkovsky+10b},
and keystroke dynamics \citep{plank16a}.
Conversely, these externalized %
signals
help us gain insight into constituency representations.
We consider two such data sources:

\paragraph{a) Answer fragments}
When questions are answered with fragments instead of full sentences,
those fragments tend to form constituents.
This phenomenon corresponds to a well-established constituency test in the linguistics literature \citep[pg. 98, inter alia]{carnie12}.

\paragraph{b) Webpage hyperlinks}
Since a hyperlink is a pointer to another location or action (e.g., {\sf mailto:} links), anchor text often represents a conceptual unit related to the link destination.
Indeed, \citet{spitkovsky+10a} first give empirical evidence that around half of the anchor text instances in their data respects constituent boundaries
and \citet{sogaard17a} demonstrates that hyperlink data can help boost chunking accuracy in a multi-task learning setup.

Both types of data have been considered in previous work on dependency-grammar induction \citep{spitkovsky+10a,naseem-barzilay11},
and in this work, we explore their efficacy for learning constituency structures.

For answer fragments, we use \posscite{he+15b} question-answering-driven semantic role labeling
(QA-SRL) dataset,
where annotators answer \emph{wh}-questions regarding predicates
in sentences drawn from the Wall Street Jounal (WSJ) section
of the Penn Treebank (PTB; \citealp{marcus+93}).
For hyperlinks, we used a $1\%$ sample of 2020-05-01 English Wikipedia, retaining only within-Wikipedia links.\footnote{
See Appendix~\ref{app:data} for details.
}

We compare our extracted naturally-occurring bracketings with
the reference phrase-structure annotations:\footnote{
For ``ground-truth'' structures in the Wikipedia data,
we apply a
state-of-the-art PTB-trained constituency parser \citep{kitaev+19}.
}
Table~\ref{tab:stats} gives relevant statistics.
Our results re-affirm \posscite{spitkovsky+10a} finding that a large proportion of hyperlinks
coincide with syntactic constituents.
We also find that
$22.4\%$/$35.8\%$ of the natural bracketings are single-word spans,
which cannot facilitate parsing decisions,
while $11.8\%$/$5.3\%$ of QA-SRL/Wikipedia spans actually conflict with the reference trees and can thus potentially harm training.
The QA-SRL data seems more promising for inducing better-quality syntactic structures,
as there are more bracketings available across a diverse set of constituent types.

\section{Parsing Model}
\label{sec:model}

\paragraph{Preliminaries}
The inputs to our learning algorithm are tuples $(w, B)$,
where
$w=w_1,\ldots,w_n$
is a length-$n$ sentence
and $B=\{(b_k,e_k)\}$
is a set of naturally-occurring bracketings,
denoted by the beginning and ending indices $b_k$ and $e_k$ into the sentence $w$.
As a first step, we extract BERT-based contextualized word representations \citep{devlin+19}
to associate each token $w_i$ with a vector $\bivec{x}_i$.\footnote{
The use of pre-trained language models can mitigate the fact that our distant supervision data are either out-of-domain (Wikipedia) or small in size (QA-SRL).
}
See Appendix~\ref{app:implementation} for details.

\paragraph{Scoring Spans}
Based on the $\bivec{x}_i$ vectors,
we assign a score $s_{ij}$ to each candidate span $(i, j)$ in the sentence
indicating its appropriateness as a constituent in the output structure.
We adopt a biaffine scoring function \citep{dozat-manning17}:
\begin{equation*}
s_{ij} = [\bivec{l}_{i};1]^T W [\bivec{r}_{j};1],
\end{equation*}
where $[\bivec{v};1]$ appends $1$ to the end of vector $\bivec{v}$,
and
\begin{equation*}
\bivec{l}_i = \text{MLP}^{\text{left}}(\bivec{x}_i)\quad\text{and}\quad \bivec{r}_j = \text{MLP}^{\text{right}}(\bivec{x}_j)
\end{equation*}
are the outputs of multi-layer perceptrons (MLPs)
that take the vectors at span boundaries as inputs.\footnote{
This is inspired by  span-based supervised constituency-parsing methods \cite{stern+17a}, which in turn was based on \citet{wang-chang16}. These papers look at the difference vectors between two boundary points, while our scoring function directly uses the vectors at the boundaries (which is more expressive than only using difference vectors).
}

\paragraph{Decoding}
We define the score $s(y)$ of a binary-branching constituency tree $y$ to be the sum of scores of its spans.
The best scoring tree among all valid trees $\mathcal{Y}$ can be found
using the CKY algorithm \citep{cocke69,kasami65,younger67}.

\paragraph{Learning}
Large-margin training \citep{taskar+05} is a typical choice
for \emph{supervised} training of constituency parsers.
It defines the following loss function to encourage a large margin of at least $\Delta(y,y^*)$ between
the gold tree $y^*$ and any predicted tree $y$:
\begin{equation*}
    l=\max_{y\in \mathcal{Y}}\left[s(y) + \Delta(y, y^*)\right] - s(y^*),
\end{equation*}
where $\Delta(y,y^*)$ is a distance measure between $y$ and $y^*$.
We can reuse the CKY decoder for cost-augmented inference
when the distance decomposes into individual spans with some function $c$:
\begin{equation*}
\textstyle
\Delta(y, y^*) = \sum_{\text{span }(i, j) \text{ in } y} c(i, j, y^*).
\end{equation*}

In our setting, we do not have access to the gold-standard $y^*$,
but instead we have a set of bracketings $\tilde{y}$.
The scoring $s(\tilde{y})$ is not meaningful
since $\tilde{y}$ is not a complete tree,
so we adopt structured ramp loss \citep{do+08,gimpel-smith12} and define
\begin{align*}
    l=&\left(\max_{y\in \mathcal{Y}}\left[s(y) + \Delta(y, \tilde{y})\right] - s(\tilde{y})\right)\\
    &+ \left(s(\tilde{y})- \max_{y\in \mathcal{Y}}\left[s(y) - \Delta(y, \tilde{y})\right]\right)\\
    =&\max_{y\in \mathcal{Y}}\left[s(y) + \Delta(y, \tilde{y})\right]\\
    &- \max_{y\in \mathcal{Y}}\left[s(y) - \Delta(y, \tilde{y})\right],
\end{align*}
using a combination of cost-augmented and cost-diminished inference.
This loss function can be understood as a sum of a convex and a concave large margin loss \citep{collobert+06},
canceling out the term for directly scoring the gold-standard tree.
We consider two methods for incorporating the partial bracketings into the cost functions:
\begin{align*}
    c_\text{loose}&(i,j,\tilde{y}) = \mathbbm{1}(\text{span }(i,j) \text{ conflicts with } \tilde{y})\\
    c_\text{strict}&(i,j,\tilde{y}) = \mathbbm{1}(\text{span }(i,j) \text{ not in } \tilde{y}),
\end{align*}
where $\mathbbm{1}$ is an indicator function.
$c_\text{loose}$ is more lenient than $c_\text{strict}$ as it does not penalize spans that do not conflict with $\tilde{y}$.
Both cost definitions promote structures containing bracketings in $\tilde{y}$.\footnote{
One may also consider a linear interpolation of $c_\text{loose}$ and $c_\text{strict}$,
but that would introduce an additional hyper-parameter.
}
In the supervised setting where $\tilde{y}$ refers to
a fully-annotated tree $y^*$ without conflicting span boundaries,
$c_\text{strict}$ is equal to $c_\text{loose}$
and the resulting $\Delta(y,y^*)$ cost functions
both correspond to the Hamming distance between $y$ and $y^*$.

\section{Experiments and Results}
\label{sec:exp}

\begin{table}[t]
    \centering
    \small
    \begin{tabular}{l@{\hspace{4pt}}c@{\hspace{4pt}}cc}
    \toprule
    Model          & PLM & Mean        & Max       \\
    \midrule
    Random Trees   & &  $19.2$      & $19.5$     \\
    Left Branching & & \multicolumn{2}{c}{$8.7$} \\
    Right Branching & & \multicolumn{2}{c}{$39.5$} \\
    \midrule
    Upper bound & & \multicolumn{2}{c}{$84.3$} \\
    \midrule
    URNNG \citep{kim+19}      & & ---    & $45.4$ \\
    PRPN \citep{shen+18a}     & & $47.3$ & $47.9$ \\
    ON \citep{shen+19}        & & $48.1$ & $50.0$ \\
    DIORA \citep{drozdov+19a} & $\circ$ & ---    & $58.9$ \\
    CPCFG \citep{kim+19a}     & & $55.2$ & $60.1$ \\
    S-DIORA \citep{drozdov+20}     & $\circ$ & $57.6$ & $64.0$ \\
    Constituency Tests \citep{cao+20}     & $\bullet$ & $62.8$ & $65.9$ \\
    \quad +URNNG \citep{cao+20}     & $\bullet$ & $67.9$ & $\mathbf{71.3}$ \\
    \underline{\emph{This work:}} & &    & \\
    NOB\textsubscript{QA-SRL}, $c_\text{loose}$
                              & $\bullet$ & $64.5$ & $65.2$ \\
    NOB\textsubscript{QA-SRL}, $c_\text{strict}$
                              & $\bullet$ & $\mathbf{68.9}$ & $70.0$ \\
    NOB\textsubscript{Wikipedia}, $c_\text{loose}$
                              & $\bullet$ & $58.2$ & $63.0$ \\
    NOB\textsubscript{Wikipedia}, $c_\text{strict}$
                              & $\bullet$ & $56.1$ & $57.0$ \\

    \bottomrule
    \end{tabular}
    \caption{Sentence-level unlabeled F1 scores ($\%$) on the WSJ test set.
    $\bullet$ in the PLM column denotes the use of context-sensitive pre-trained language models;
    $\circ$
    uses context-insensitive embedders from PLMs.
    Methods producing only binary-branching structures
    (including everything in this table)
    have an upperbound of $84.3\%$ F1 score,
    since the gold trees can be non-binary.
    }
    \label{tab:wsj}
\end{table}

\paragraph{Data and Implementation}
We evaluate on the PTB \citep{marcus+93} with the standard splits (section 23 as the test set).
QA-SRL contains $1{,}241$ sentences drawn from the training split (sections 02-21) of the PTB.
For Wikipedia, we use a sample of $332{,}079$ sentences
that are within $100$ tokens long and contain multi-token internal hyperlinks.
We fine-tune the pretrained BERT\textsubscript{base} features
with a fixed number of mini-batch updates
and report results based on five random runs for each setting.
See Appendix~\ref{app:implementation} for detailed hyper-parameter settings and optimization procedures.

\paragraph{Evaluation}
We follow the evaluation setting of \citet{kim+19a}.
More specifically,
we discard punctuation and trivial spans (single-word and full-sentence spans) during evaluation
and report sentence-level F1 scores as our main metrics.

\begin{table}[t]
    \centering
    \small
    \begin{tabular}{l|c|cc|cc}
    \toprule
    \multirow{2}{*}{\begin{tabular}[c]{@{}l@{}}Const.\\ Type\end{tabular}}
     & \multirow{2}{*}{\begin{tabular}[c]{@{}l@{}}\citeauthor{cao+20}\\ (\citeyear{cao+20})\end{tabular}}
      & \multicolumn{2}{c|}{NOB\textsubscript{QA-SRL}} & \multicolumn{2}{c}{NOB\textsubscript{Wikipedia}} \\
    &  & $c_\text{loose}$ & $c_\text{strict}$ &  $c_\text{loose}$ & $c_\text{strict}$ \\
    \midrule
    SBAR & $85.3$ & $\mathbf{89.0}$ & $87.7$ & $66.7$ & $48.3$ \\
    NP   & $84.3$ & $\mathbf{85.5}$ & $85.2$ & $71.8$ & $70.3$ \\
    VP   & $\mathbf{80.8}$ & $52.3$ & $70.9$ & $62.4$ & $49.6$ \\
    PP   & $84.4$ & $83.5$ & $\mathbf{86.5}$ & $67.7$ & $74.8$ \\
    ADJP & $55.6$ & $58.1$ & $57.3$ & $\mathbf{62.7}$ & $60.1$ \\
    ADVP & $54.6$ & $\mathbf{76.9}$ & $75.3$ & $66.4$ & $63.9$ \\
    \bottomrule
    \end{tabular}
    \caption{Average recall (\%) per consituent type.}
    \label{tab:per-type}
\end{table}

\paragraph{Results}
Table~\ref{tab:wsj} shows the evaluation results of our models trained on naturally-occurring bracketings (NOB);
Table~\ref{tab:per-type} breaks down the recall ratios for each constituent type.
Our distantly-supervised models trained on QA-SRL
are competitive with the state-of-the-art unsupervised results.
When comparing our models with \citet{cao+20},
we obtain higher recalls on most constituent types except for VPs.
Interestingly, QA-SRL data prefers $c_\text{strict}$,
while $c_\text{loose}$ gives better F1 score on Wikipedia;
this correlates with the fact that QA-SRL has more bracketings per sentence (Table~\ref{tab:stats}).
Finally, our Wikipedia data has a larger relative percentage of ADJP bracketings,
which explains the higher ADJP recall of the models trained on Wikipedia,
despite their lower overall recalls.

\section{Related Work}
\label{sec:related}

\paragraph{Unsupervised Parsing}
Our distantly-supervised setting is similar to unsupervised in the sense that it does not require syntactic annotations.
Typically, lack of annotations implies that unsupervised parsers
induce grammar from a raw stream of lexical or part-of-speech tokens \citep{clark01,klein05}
along with carefully designed inductive biases on parameter priors \citep{liang+07,wang-blunsom13},
language universals \citep{naseem+10,martinezalonso+17},
cross-linguistic \citep{snyder+09,berg-kirkpatrick-klein10,cohen-smith09,han+19} and cross-modal \citep{shi+19} signals,
structural constraints \citep{gillenwater+10,noji+16,jin+18}, etc.
The models are usually generative and learn from (re)constructing sentences based on induced structures \citep{shen+18a,shen+19,drozdov+19a,kim+19a,kim+19}.
Alternatively, one may use reinforcement learning to induce syntactic structures using rewards defined by end tasks \citep{yogatama+17,choi+18,havrylov+19}.
Our method is related
to learning from constituency tests \citep{cao+20},
but
our use of bracketing data
permits discriminative parsing models,
which focus directly on the syntactic objective.

\paragraph{Learning from Partial Annotations}
Full syntactic annotations are costly to obtain,
so the alternative solution of training parsers from partially-annotated data has attracted considerable research attention,
especially within the context of active learning for dependency parsing
\citep{sassano05,sassano-kurohashi10,mirroshandel-nasr11,flannery+11,flannery-mori15,li+16k,zhang+17b}
and grammar induction for constituency parsing
\citep{pereira-schabes92,hwa99,riezler+02}.
These works typically require expert annotators to generate gold-standard, though partial, annotations.
In contrast, our work considers the setting and the challenge of learning from noisy bracketing data,
which is more comparable to \citet{spreyer-kuhn09} and \citet{spreyer+10} on transfer learning for dependency parsing.

\section{Conclusion and Future Work}
\label{sec:conclusion}

We argue that naturally-occurring bracketings are a rich resource for inducing syntactic structures.
They reflect human judgment of what constitutes a phrase and what does not.
More importantly,
they require low annotation expertise and effort;
for example,
webpage hyperlinks can be extracted essentially for free.
Empirically,
our models trained on QA-SRL and Wikipedia bracketings
achieve competitive results with the state of the art
on unsupervised constituency parsing.

Structural probes have been successful in extracting syntactic knowledge
from frozen-weight pre-trained language models (e.g., \citealp{hewitt-manning19}),
but they still require direct syntactic supervision.
Our work shows that it is also feasible to
retrieve constituency trees from BERT-based models
using distant supervision data.

Our models are limited to the unlabeled setting,
and we leave it to future work to automatically cluster the naturally-occurring bracketings and to induce phrase labels.
Our work also points to potential applications in (semi-)supervised settings including active learning and domain adaptation  \citep{joshi+18}.
Future work can also consider other naturally-occurring bracketings
induced from sources such as speech production, reading behavior, etc.

\section*{Acknowledgements}

We thank the anonymous reviewers for their constructive reviews.
This work was supported in part by a Bloomberg Data Science Ph.D. Fellowship to Tianze Shi and a gift from Bloomberg to Lillian Lee.

\bibliographystyle{acl_natbib}
\bibliography{ref}

\clearpage
\appendix

\section{Data}
\label{app:data}

\subsection{QA-SRL}

\posscite{he+15b} question-answering-driven semantic role labeling dataset (QA-SRL) contains question-answer pairs for $1{,}241$ sentences drawn originally from the training sections of the Penn Treebank (PTB; \citealp{marcus+93}).
The questions are generated by templates
that ask about semantic arguments for all the predicates in a given sentence.
Recorded human responses to the questions typically correspond to spans in the sentence.
Each question can have multiple answers.

For all question-answer pairs,
we first map the answers to consecutive spans in the corresponding sentences.
We keep all exact matches when the answer text appears multiple times in the sentence,
and we discard any answers that cannot be mapped to a consecutive span in the sentence.

\subsection{Wikipedia}

We randomly sample $1\%$ of the articles from the 2020-05-01 snapshot of English Wikipedia\footnote{
\url{https://dumps.wikimedia.org/enwiki/}
}.
We then split the documents into sentences and tokenize with spaCy.\footnote{\url{https://spacy.io}}
This step leads to $926{,}077$ sentences,
as reported in Table~\ref{tab:stats}.
For ground-truth parse trees, we parse the sentences with \posscite{kitaev+19}  state-of-the-art constituency parser trained on the PTB.
For all the internal hyperlinks in the documents, where there is a hyperlink-tokenization mismatch, we retrieve
the
 smallest span of tokens that covers the hyperlink.
To construct the training set in our main experiments, we filter out sentences longer than $100$ tokens and sentences without any multiple-token internal hyperlinks.
These pre-processing procedures produce $332{,}079$ training sentences.

\section{Implementation Details}
\label{app:implementation}

\paragraph{Feature Extractor}
We use the pretrained BERT\textsubscript{base} model as our feature extractor.\footnote{Pytorch interface of the model is provided by \url{https://github.com/huggingface/transformers}.}
For each word in the sentence, we tokenize it with BERT's WordPiece tokenizer,
and we take the BERT vector of the last token at the final BERT hidden layer as representation for each word.
The feature extractor is fine-tuned along with model training.

\paragraph{Span Scoring}
MLP\textsuperscript{left} and MLP\textsuperscript{right} are single-layer MLPs: they both consist of a linear layer projecting BERT representations to $256$-dimensional vectors, followed by a leaky ReLU activation function \citep{maas+13}.
The constituent scoring component has parameter $W\in \mathbb{R}^{257\times 257}$.
All the parameters are randomly initialized \citep{glorot-bengio10}.

\paragraph{Training and Optimization}
We optimize the neural networks using the Adam optimizer \citep{kingma-ba15}
with $\beta_1=0.9$, $\beta_2=0.999$ and $\epsilon=1\times 10^{-12}$.
For each batch, we sample $8$ sentences from the training set and average the loss collected for each sentence.
The gradients are clipped at $1.0$ before back propagation.
The learning rate linearly increases from zero to $1 \times 10^{-5}$ in $2{,}000$ training steps.
After warmup, we keep training the model until we reach $20{,}000$ training steps.
We do not perform early stopping,
since in the unsupervised parsing setting, we do not look at validation accuracies until we finish training.
We leave it as future work to explore other model selection strategies.

\paragraph{Hyperparameter Selection}
We use the default recommended $\beta_1$, $\beta_2$, and $\epsilon$ values for the Adam optimizer,
and we use a typical
fine-tuning learning rate
for the pre-trained BERT model \citep{devlin+19}.
The number of training steps is based on our preliminary observation of the convergence of the training loss,
and the batch size is limited by our computating hardware.
We fix the initial values we set for the size of the biaffine matrix ($257\times 257$) and the number of warmup steps ($2{,}000$) throughout our experiments.
A better hyperparameter selection strategy may lead to improved results.

\paragraph{Speed}

For a length-$n$ sentence, the time complexity for the CKY decoder is $O(n^3)$.
On a RTX 2080 GPU, our model parses $409$ sentences per second on average
and the training process for each model finishes within $2$ hours.

\end{document}